\pdfoutput=1
\documentclass[letterpaper, 10 pt, journal, twoside]{IEEEtran}  

\IEEEoverridecommandlockouts                              

\usepackage{amsmath}
\usepackage[linesnumbered,algoruled,boxed,lined]{algorithm2e}
\usepackage[noend]{algpseudocode}
\usepackage{amssymb}
\usepackage{amsmath}
\usepackage{hyperref}
\usepackage{graphicx}
\usepackage{caption}
\graphicspath{ {./images/} }
\usepackage{color}
\usepackage[table]{xcolor}
\usepackage{url}
\usepackage{leftidx}
\usepackage[noadjust]{cite}
\usepackage{filecontents}
\usepackage{booktabs}
\usepackage{soul}
\usepackage{array}
\usepackage{ulem}
\usepackage{tikz}
\usepackage{boldline}

\usepackage{graphics} 
\usepackage{booktabs}
\usepackage{subfigure}
\usepackage{multirow}
\usepackage{threeparttable}
\usepackage{overpic}
\usepackage{makecell}

\usetikzlibrary{positioning}
\usepackage[top=0.75in, bottom=0.6in, left=0.67in, right=0.67in]{geometry}

\usepackage{microtype}
\usepackage{tabularray}
\usepackage{gensymb}

\title{\LARGE ERRA: An Embodied Representation and Reasoning Architecture for Long-horizon Language-conditioned Manipulation Tasks}

\author{Chao Zhao\textsuperscript{*}, Shuai Yuan\textsuperscript{*}, Chunli Jiang, Junhao Cai, \\ Hongyu Yu, Michael Yu Wang, and Qifeng Chen 
\thanks{Manuscript received: December, 5, 2022; Revised February, 28, 2023; Accepted March, 29, 2023. }

\thanks{This paper was recommended for publication by Editor Hong Liu upon evaluation of the Associate Editor and Reviewers’ comments.}

\thanks{\textsuperscript{*}Authors with equal contribution. C. Zhao, S. Yuan, C. Jiang, J. Cai, H. Yu, and Q. Chen are with The Hong Kong University of Science and Technology, Clear Water Bay, Hong Kong  {\tt\footnotesize \{czhaobb, syuanaf, cjiangab, jcaiaq\}@connect.ust.hk} and {\tt\footnotesize \{hongyuyu, cqf\}@ust.hk}. J. Cai, H. Yu, and M. Wang are also with HKUST Shenzhen-Hong Kong Collaborative Innovation Research Institute, Futian, Shenzhen. M. Wang is with Monash University {\tt\footnotesize michael.y.wang@monash.edu}}

\thanks{Digital Object Identifier (DOI): see the top of this page.} 
}

\begin{document}

\maketitle
\begin{abstract}
This letter introduces ERRA, an embodied learning architecture that enables robots to jointly obtain three fundamental capabilities (reasoning, planning, and interaction) for solving long-horizon language-conditioned manipulation tasks. ERRA is based on tightly-coupled probabilistic inferences at two granularity levels. Coarse-resolution inference is formulated as sequence generation through a large language model, which infers action language from natural language instruction and environment state. The robot then zooms to the fine-resolution inference part to perform the concrete action corresponding to the action language. Fine-resolution inference is constructed as a Markov decision process, which takes action language and environmental sensing as observations and outputs the action. The results of action execution in environments provide feedback for subsequent coarse-resolution reasoning. Such coarse-to-fine inference allows the robot to decompose and achieve long-horizon tasks interactively. In extensive experiments, we show that ERRA can complete various long-horizon manipulation tasks specified by abstract language instructions. We also demonstrate successful generalization to the novel but similar natural language instructions.

\end{abstract}

\section{Introduction}\label{intro}

If robots are to be widely deployed in workplaces, hospitals, and our homes to assist us, they must understand our needs, discover the underlying causal relations of environments, and interact with the environment appropriately. An example is the case of long-horizon manipulation tasks specified by natural language. For example, when humans hear a request such as ``Please put the cosmetic in the drawer'', we can simultaneously understand the sentence's semantics and observe the surroundings to determine whether we need to  ``open the drawer'' first or ``grasp the cosmetic.'' We then observe the outcomes of attempted concrete action and plan next. In addition, we can take corrective measures from failure cases (e.g., cosmetic slips from our hands). To operate in our world, robots must replicate such abilities. 

This is the motivation for the problem tackled in this paper, which is a robot that has the following abilities: (i) reason abstract nature language instructions and plan with the causal relation of the environment, (ii) develop motor skills to interact with environments, and complete long-horizon manipulation tasks, (iii) detect failures (e.g., accidentally drop an object) and correct them (e.g., grasp the object again). Endowing robots with the combination of abilities (i)-(iii) is a grand challenge because the long-horizon manipulation tasks with abstract language instructions, for example, ``clean trash on the table,''  requires the embodied agent to have semantic knowledge and a reliable interpretation of the environment, to successfully plan and perform a long sequence of motor skills, and to know when to stop (i.e., no trash on the table). While conventional methods, such as symbolic programming or hierarchical reinforcement learning, can plan tasks, most approaches rely on carefully designed representations and analytical transition models, which limit generalization. Recently, a few studies have explored the use of pre-trained large language models (LLMs) to answer questions that require reasoning and planning through prompt design (i.e., hand-crafted text prompts) and utilizing such ability for long-horizon robot manipulation \cite{saycan, inner, huang2022language}. However, an important problem with these approaches is that there is no guarantee of what manipulation tasks LLMs can reason about and plan without trying because LLMs lack real-world experience during their original training. Furthermore, small changes in prompts can deteriorate the performance of LLMs, making finding appropriate prompts time-consuming.

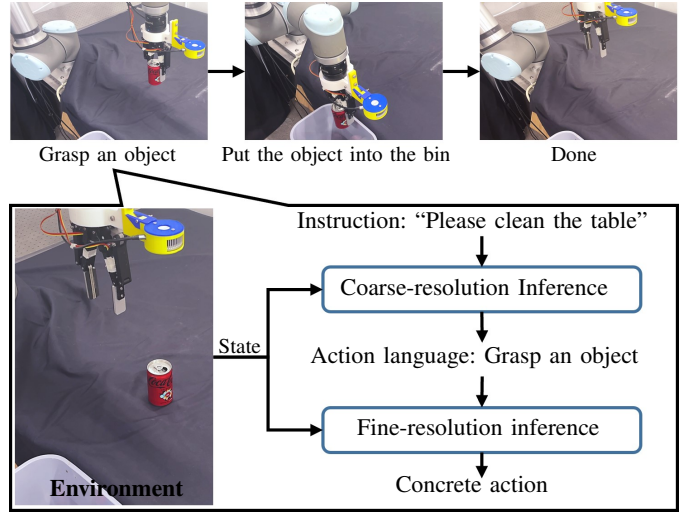
\begin{figure}[!t]
    \centering
    \begin{overpic}[width=\linewidth]{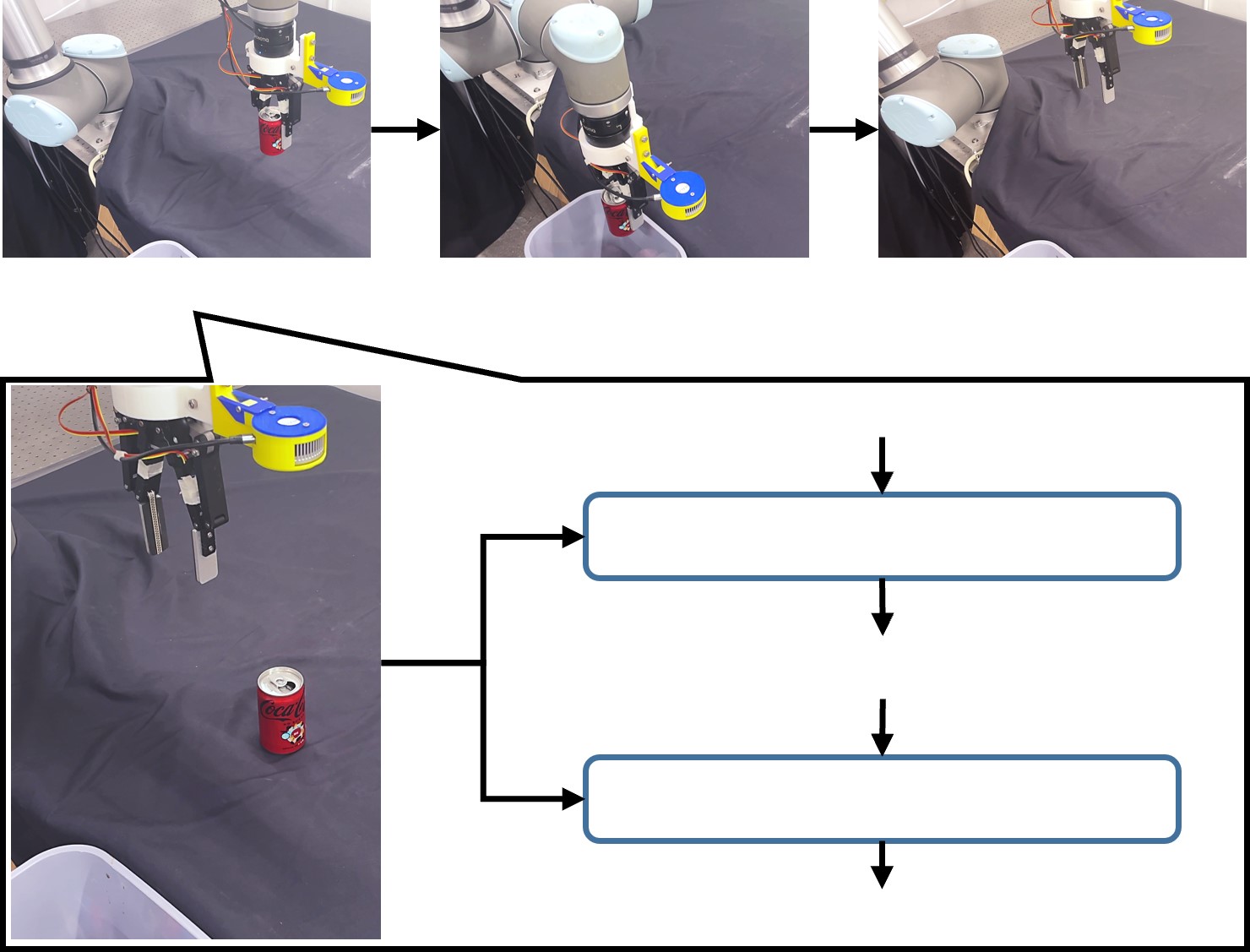}
    \put(6,2.25) {\small {\textbf{Environment}}}
    \put(31.25,23.7) {\footnotesize {State}}
    \put(43,42.4) {\small {Instruction: ``Please clean the table''}}
    \put(49.5,32.4) {\small {Coarse-resolution Inference}}
    \put(45,22.25) {\small {Action language: Grasp an object}}
    \put(52,11.4) {\small {Fine-resolution inference}}
    \put(57.75,2.8) {\small {Concrete action}}
    \put(3,52.5)  { \footnotesize {Grasp an object}}
    \put(31.75,52.5) {\footnotesize	 {Put the object into the bin}}
    \put(81,52.5) {\footnotesize	 {Done}}
    \end{overpic}
    \caption{
    \textbf{ERRA Overview.} The image sequence at the top shows the action language and execution process to complete a task with the instruction ``Please clean the table,'' utilizing ERRA. Given a language instruction, the coarse-resolution inference produces the next step represented by action language, according to the environment state. The action language and state are then processed by fine-resolution inference, which outputs the concrete action to interact with the environment. 
    }
    \label{fig:fg0}
\vspace{-0.6cm}
\end{figure}

To address the above problems and endow robots with abilities (i)-(iii), we propose the ERRA framework based on tightly-coupled probabilistic inferences at two levels of granularity, coarse and fine. An overview of ERRA is shown in Fig. \ref{fig:fg0}. The coarse-resolution inference focuses on high-level reasoning and planning (i.e., what to do in the next step?), and the fine-resolution inference focuses on learning concrete actions (i.e., how to do it?). The results of executing concrete actions in the environment provide feedback for subsequent coarse-resolution inferences. Such coarse-to-fine inferences are invoked repeatedly to decompose long-horizon manipulation tasks as a sequence of concrete actions. Coarse-resolution inference is built on a pre-trained large language model for generating the action language. Motor skills in fine-resolution inference are learned through reinforcement learning (RL) under self-supervision. 

The primary contribution of this work is to suggest a new approach, ERRA, that allows an embodied agent to acquire reasoning, planning, and interaction abilities for solving long-horizon manipulation tasks specified by natural language. Extensive experiments show that ERRA is capable of understanding the semantics in abstract language instructions, reasoning in environments with rich functional relationships between objects, and providing motor skills to complete long-horizon manipulation tasks. We also show that ERRA allows the robot to recover from failure cases and adapt to environmental changes in the real world, significantly improving the robustness of robots in dynamic environments. 

\section{Related Work}

\textbf{Task and Motion Planning.} In robotics, task and motion planning \cite{akbari2019knowledge} is capable of solving long-horizon (i.e., multi-step) manipulation tasks. Traditional methods rely on symbolic planning \cite{sy1} or optimization \cite{op1} in abstract or symbolic spaces. However, most approaches require manually defined representation spaces and environment kinematics models, which are usually domain-specific and lack generalization ability. More recently, LLMs have demonstrated dawning properties on reasoning and planning under appropriate conditions (e.g., language prompts) \cite{llm1, llm2, llm3, llm4}. Several works \cite{huang2022language} have studied using LLMs to plan robot manipulation tasks. SayCan \cite{saycan} uses LLM to infer the entire plans of the manipulation task and estimate the feasibility of each step using a model of action affordance. However, these methods assume that the execution of each planned motor skill is faultless, making them not robust to intermediate failures in task execution. In this aspect, \cite{inner} introduces additional modules to incorporate human and environmental feedback to improve the completion of tasks. While prior works have investigated how LLMs plan via prompt design, the ability of LLMs is agnostic, requiring time and human effort to design and experiment with different prompts. We introduce the prompt tuning method \cite{prefix}, enabling the LLM to be a reasoner and planner without using design prompts.

\textbf{Learning Language-Conditioned Manipulation.} Natural language provides a human-interactive interface to link humans to robots, which is important for deploying robots in our lives. Many studies \cite{lcmf-1, lcmf-2, lcmf-3, lcmf-4, lcmf-5} have explored how robots follow language instructions, in which robots are required to complete tasks specified by the language. Some studies \cite{lcm-1-bcz, lcm-2, lcm-3} have learned language-conditioned behaviors through imitation learning. For example, \cite{lcm-e1} learns a direct mapping from images and natural language instructions to actions using a Transformer network. \cite{lcm-e2} uses an offline robotics dataset with crowdsourced natural language labels to learn a range of vision-based manipulation tasks. Most of these works focus on learning short-horizon manipulation tasks such as grasping or in-hand manipulation. In contrast, ERRA can understand instructions with abstract semantics and achieve long-horizon tasks by leveraging LLMs' semantic knowledge to interpret instructions and plan tasks. 

\textbf{Reinforcement Learning for Manipulation.} Reinforcement learning combined with deep learning has recently made extensive progress in learning skills in different domains, such as beyond human experts at the games of Go \cite{gozero} and Atari \cite{atari}. In robotic manipulation, reinforcement learning offers the robot a way to acquire various manipulation skills through self-exploration \cite{rl2,rl3,rl4}. However, most studies focus on learning narrow and individual tasks. Some works achieve long-horizon task planning by hierarchical reinforcement learning \cite{hrl1,hrl2}, which requires manual task-level design and lacks generalization ability. In our work, ERRA leverages reinforcement learning to acquire low-level motor skills in the simulation and cooperates with the coarse-resolution inference module to perform long-horizon manipulation tasks.

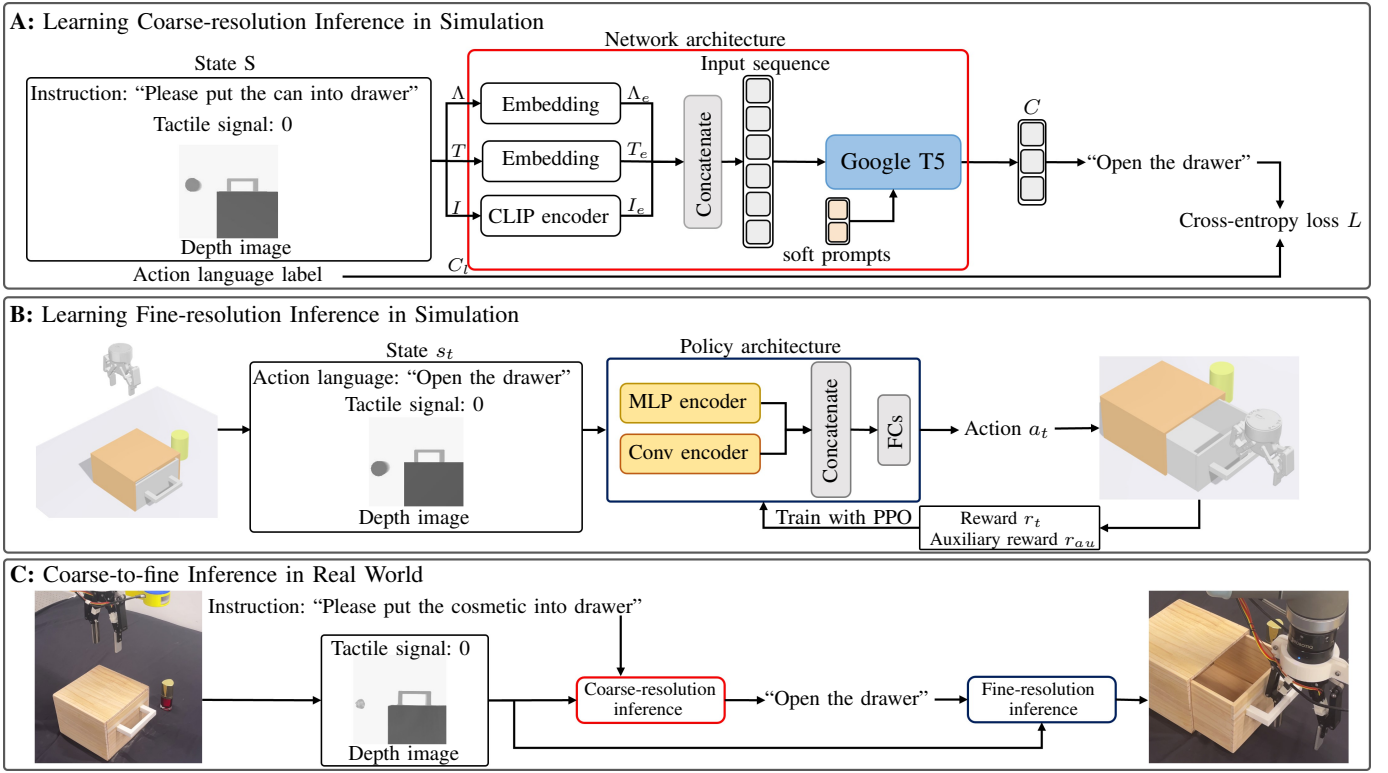
\begin{figure*}[!htbp]
    \centering
    \begin{overpic}[width=\linewidth]{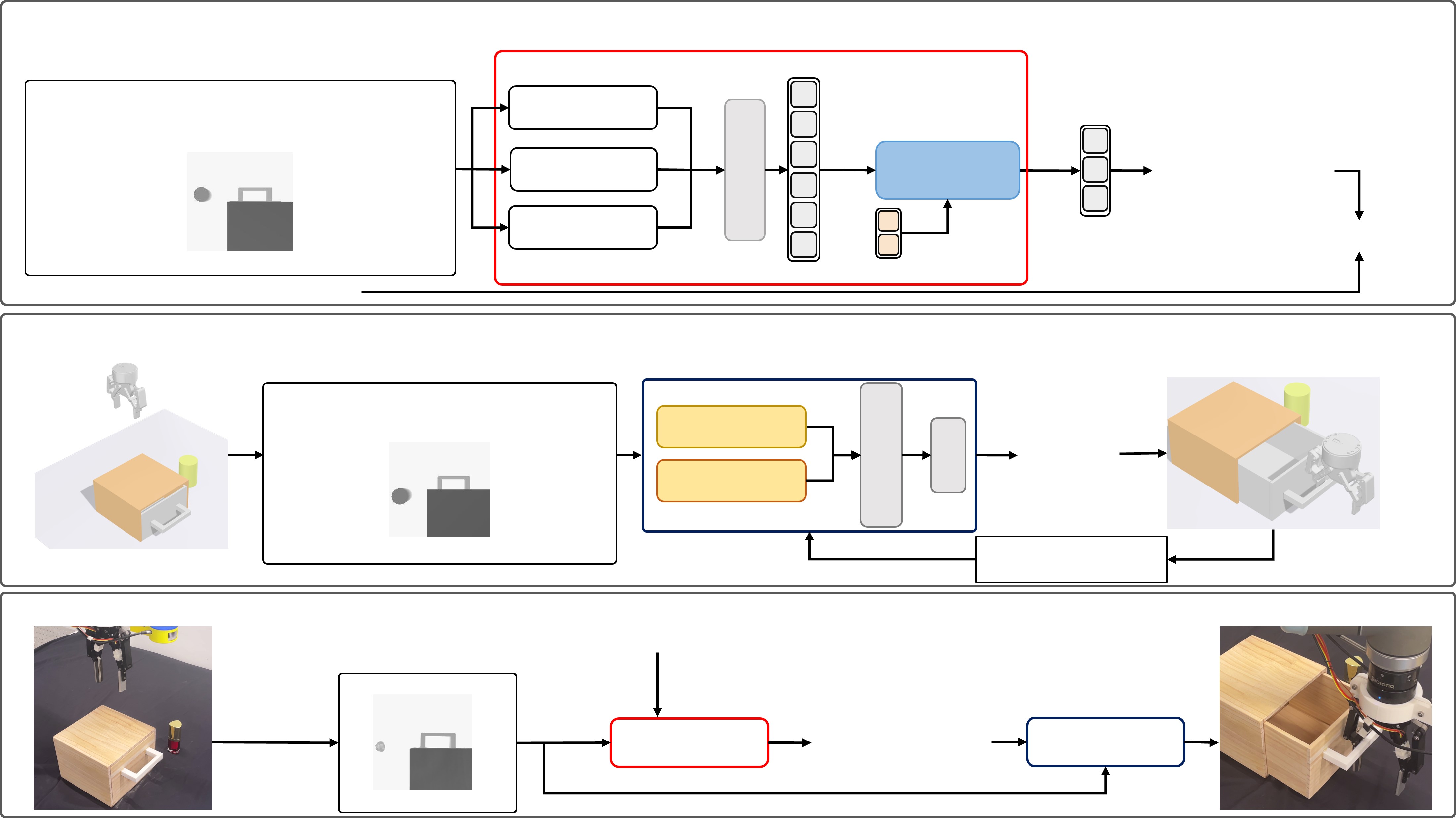}
    \put(0.5,54.25) {\small {\textbf{A:} Learning Coarse-resolution Inference in Simulation}}
    \put(14,51.25) {\footnotesize {State S}}
    \put(2,49) {\footnotesize {Instruction: ``Please put the can into drawer''}}
    \put(11,46.75) {\footnotesize {Tactile signal: 0}}
    \put(13,37.75) {\footnotesize {Depth image}}
    \put(9.5,35.75) {\footnotesize {Action language label}}

    \put(32.75,49.1) {\scriptsize {$\Lambda$}}
    \put(32.75,44.85) {\scriptsize {$T$}}
    \put(32.75,40.7) {\scriptsize {$I$}}
    \put(32.5,36.5) {\scriptsize {$C_l$}}

    \put(45.6,49.1) {\scriptsize {$\Lambda_e$}}
    \put(45.6,45) {\scriptsize {$T_e$}}
    \put(45.6,40.9) {\scriptsize {$I_e$}}

    \put(50, 40.5){ {\rotatebox{90}{\footnotesize  Concatenate}}}
    \put(59.25, 21){ {\rotatebox{90}{\footnotesize  Concatenate}}}
    \put(64, 23.5){ {\rotatebox{90}{\footnotesize  FCs}}}
    
    \put(36.5,48.25) {\footnotesize {Embedding}}
    \put(36.5,44.25) {\footnotesize {Embedding}}
    \put(35.5,40) {\footnotesize {CLIP encoder}}
    \put(44,53) {\footnotesize {Network architecture}}
    \put(51,51.25) {\footnotesize {Input sequence}}
    \put(61.25,44) {\small {Google T5}}
    \put(57,37.25) {\footnotesize {soft prompts}}
    \put(74.6,48) {\footnotesize {$C$}}
    \put(79.25,44) {\footnotesize {``Open the drawer''}}
    \put(86,39.75) {\footnotesize {Cross-entropy loss $L$}}

    \put(0.5,32.75) {\small {\textbf{B:} Learning Fine-resolution Inference in Simulation}}
    \put(28,30.25) {\footnotesize {State $s_t$}}
    \put(18.25,28.25) {\footnotesize  {Action language: ``Open the drawer''}}
    \put(25,26.25) {\footnotesize  {Tactile signal: 0}}
    \put(26,18) {\footnotesize  {Depth image}}
    \put(49.5,30.5) {\footnotesize {Policy architecture}}
    \put(45.75,26.5) {\footnotesize  {MLP encoder}}
    \put(45.75,22.75) {\footnotesize  {Conv encoder}}
    \put(56.5,18) {\footnotesize  {Train with PPO}}
    \put(70.25,24.5) {\footnotesize  {Action $a_t$}}
    \put(70,18) {\scriptsize {Reward $r_t$}}
    \put(67.75,16.5) {\scriptsize  {Auxiliary reward $r_{au}$}}
    
    \put(0.5,13.75) {\small {\textbf{C:} Coarse-to-fine Inference in Real World}}
    \put(15,11.5) {\footnotesize {Instruction: ``Please put the cosmetic into drawer''}}
    \put(25.25,0.75){\footnotesize {Depth image}}
    \put(24,8.4) {\footnotesize {Tactile signal: 0}}
    \put(55.7,4.75) {\footnotesize {``Open the drawer''}}
    \put(42.5,5.5) {\scriptsize {Coarse-resolution}}
    \put(44.65,4) {\scriptsize {inference}}

    \put(71.55,5.5) {\scriptsize {Fine-resolution}}
    \put(73.55,4) {\scriptsize {inference}}

    \end{overpic}
    \caption{\textbf{System Overview.} \textbf{A:} We generate a set of correspondences $(S, C)$ in the simulation to learn the coarse-resolution inference. State $S$ includes the instruction $\Lambda$, depth image $I$, and tactile signal $T$. The provided inputs are encoded as three vectors $(\Lambda_e, T_e, I_e)$, respectively. These vectors are concatenated and subsequently fed to Google T5. At last, the output action language $C$ and label of action language $C_l$ are used to compute loss; \textbf{B:}  To learn fine-resolution inference, we employ PPO. The RL agent takes the state $s_t$ as input and predicts the action $a_t$ for the robot execution at time step $t$. The agent then obtains rewards $r_t$ and auxiliary reward $r_{au}$ from the simulation; \textbf{C:} We deploy the ERRA in the real world. Given instruction, the coarse-resolution module infers the action language based on current observation and the tactile signal. Then the fine-resolution module predicts actions with the inputs of action language and environment state. }
    \label{fig:fg1}
\vspace{-0.6cm}
\end{figure*}

\section{Method}\label{sec:method}

In this section, we describe the architecture of ERRA, as shown in Fig. \ref{fig:fg1}. ERRA is based on two inference modules, coarse and fine. The coarse-resolution module infers an action language (e.g., grasp the apple) based on language instruction, environment state, and robot proprioception. The action language corresponds to a motor skill that the robot needs to execute. Subsequently, the fine-resolution inference module generates concrete actions for executing the motor skill, using inputs of the action language inferred by the coarse-resolution inference module and the visual and tactile information. By iteratively invoking the coarse-to-fine inference process, a task can be decomposed into simpler concrete actions and executed. The step-by-step planning and execution processes of ERRA enable feedback to be established and mitigate the challenges of reasoning and planning, resulting in effective and robust performance. In the following sections, we describe the problem formulation of the coarse and fine-resolution inferences and elucidate the supervised learning and reinforcement learning approaches we adopted to train these two inference modules in the simulation.

\subsection{Coarse-resolution Inference}\label{sec:learn_intention}

The objective of learning the coarse-resolution inference is to obtain a high-level manipulation planning strategy, as shown in Fig. \ref{fig:fg1}A. The coarse-resolution inference is formulated as a sequence-to-sequence text generation task, in which the generated action language guides fine-resolution inference to predict concrete actions performed by the robot.

\textbf{Problem Formulation:} Formally, it is a mapping $p: S \rightarrow C$, where $S = (\Lambda, I, T)$ is the input state and  $C = (c_1,c_2,...,c_m)$ is a sequence of text that represents the action language. The state $S$ consists of three parts: a language instruction $\Lambda=(\lambda_1,\lambda_2,...,\lambda_n)$, which is a sequence of text words; a depth image $I$ taken by the camera in the environment; and a tactile signal $T$ (i.e., a binary signal indicating the presence or absence of objects between fingers). The coarse-resolution inference is constructed as a neural network that predicts each word $c_i$ $\in$ $C$ given the state $S$. 

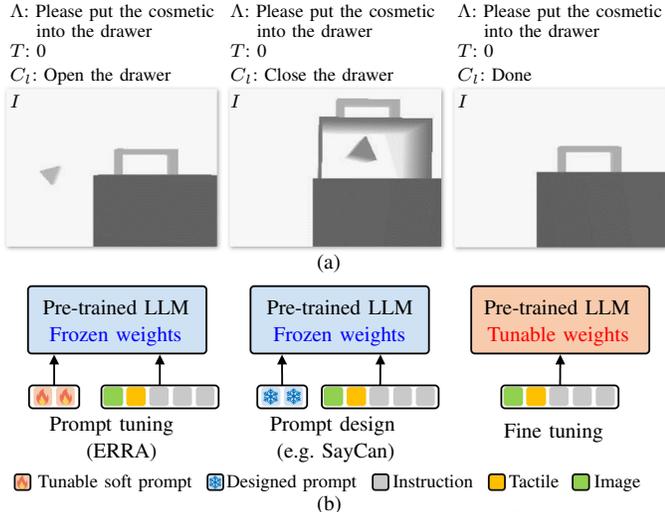
\begin{figure}[!t]
    \centering
    \begin{overpic}[width=\linewidth]{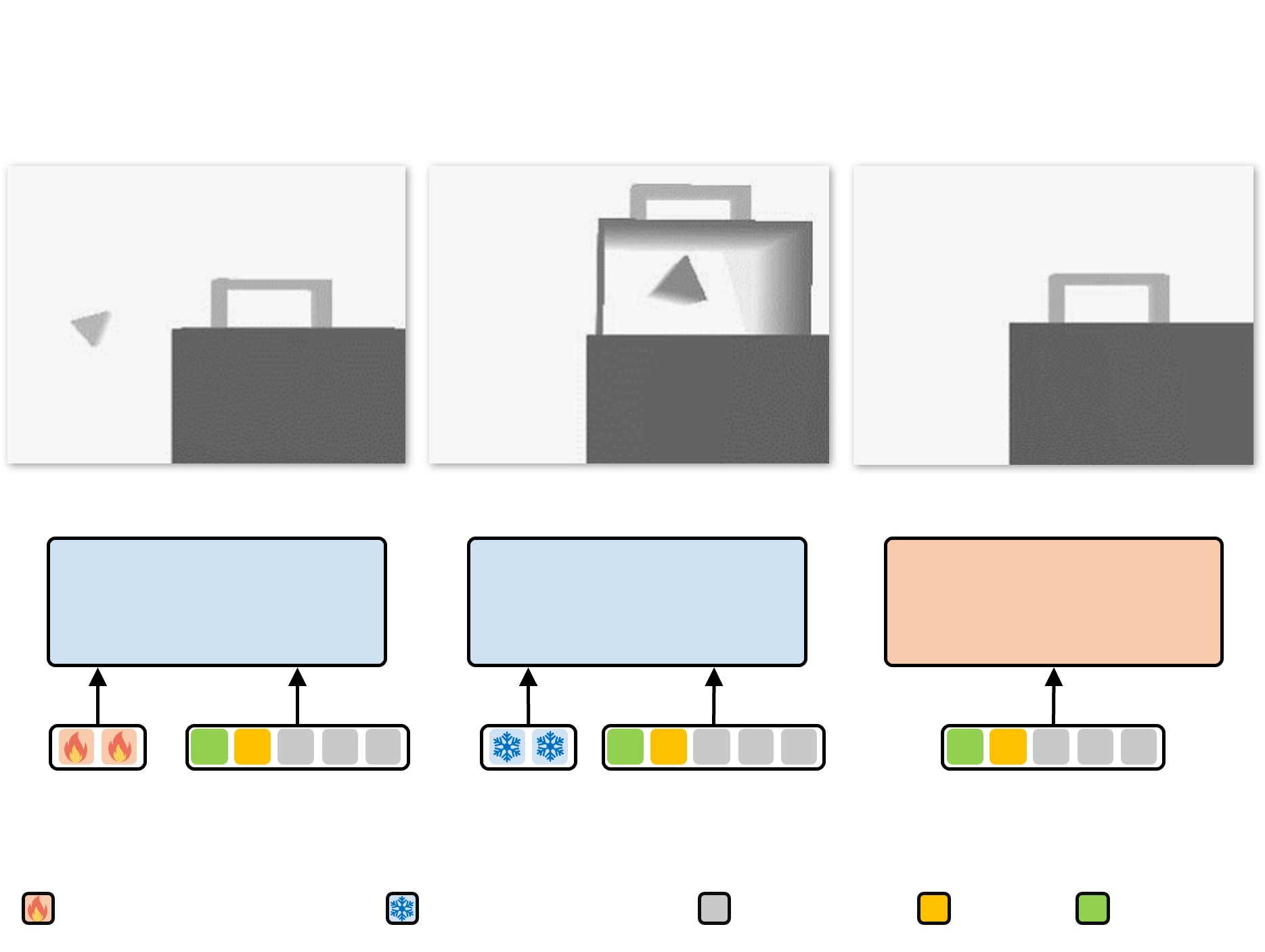}

    \put(1,73) {\scriptsize {$\Lambda$: Please put the cosmetic}}
    \put(5,70) {\scriptsize {into the drawer}}
    \put(1,67) {\scriptsize {$T$: 0}}
    \put(1,63.5) {\scriptsize {$C_l$: Open the drawer}}  
    \put(1,59.5) {\scriptsize {$I$}}  

    \put(34,73) {\scriptsize {$\Lambda$: Please put the cosmetic}}
    \put(38,70) {\scriptsize {into the drawer}}
    \put(34,67) {\scriptsize {$T$: 0}}
    \put(34,63.5) {\scriptsize {$C_l$: Close the drawer}}  
    \put(34,59.5) {\scriptsize {$I$}}  
    
    \put(68,73) {\scriptsize {$\Lambda$: Please put the cosmetic}}
    \put(72,70) {\scriptsize {into the drawer}}
    \put(68,67) {\scriptsize {$T$: 0}}
    \put(68,63.5) {\scriptsize {$C_l$: Done}} 
    \put(68,59.5) {\scriptsize {$I$}}  

    \put(47,35.25) {\footnotesize {(a)}}

    \put(7,11) {\footnotesize {Prompt tuning}}
    \put(12.5, 7) {\footnotesize {(ERRA)}}
    \put(6,28.45) {\footnotesize {Pre-trained LLM}}
    \put(7,24.45) {\footnotesize \textcolor{blue}{Frozen weights}}
    
    \put(40,11) {\footnotesize {Prompt design}}
    \put(41,7) {\footnotesize {(e.g. SayCan)}}
    \put(39.5,28.45) {\footnotesize {Pre-trained LLM}}
    \put(40.5,24.45) {\footnotesize \textcolor{blue}{Frozen weights}}

    \put(75,10) {\footnotesize {Fine tuning}}
    \put(72,28.45) {\footnotesize {Pre-trained LLM}}
    \put(72.5,24.45) {\footnotesize \textcolor{red}{Tunable weights}}

    \put(5.25,2.5) {\scriptsize {Tunable soft prompt}}
    \put(33.5,2.5) {\scriptsize {Designed prompt}}
    \put(58.35,2.5) {\scriptsize {Instruction}}
    \put(75.75,2.5) {\scriptsize {Tactile}}
    \put(88.5,2.5) {\scriptsize {Image}}
    
    \put(47,-0.9) {\footnotesize {(b)}}

    \end{overpic}
\caption{(a) Three examples of collected data, each data $d_i$ contains a state
$S = (\Lambda, I, T)$, and a label of the action language $C_l$; (b) Difference between our training and others. Instead of adjusting the parameters of LLM or using engineered prompts, our method introduces prompt tuning, which adds a small set of learnable soft prompts and shares the frozen LLM across all tasks.}
    \label{fig:fg2}
\vspace{-0.6cm}
\end{figure}

\textbf{Learning Coarse-resolution Inference:} To learn the coarse-resolution inference, we generate a synthetic dataset $D = (d_1, d_2, \dots)$. The dataset is collected in simulation leveraging the pre-programmed environments for various language-conditioned manipulation tasks. Each piece of data $d_i$ contains a corresponding relationship: at the current state $S$, what needs to do next (denoted as $C_l = (c_{l1},...,c_{lm})$). For instance, in the first example of Fig. \ref{fig:fg2}(a), the data $d_i \in D$ consists of the language instruction $\Lambda = \{Please~put~the~cosmetic~into~the~drawer\}$, the tactile signal $T = 0$, the depth image $I$, and the label of the action language $C_l = \{Open~the~drawer\}$. 


We choose Google T5 \cite{t5}, a large-scale pre-trained language model, as the backbone of the network, which provides benefits of better contextual understanding and generalization ability for language. As shown in Fig. \ref{fig:fg1}A, the model extracts the information from the language instruction, tactile signal, and image. Then, the model reasons the next action language is ``Open the drawer,'' based on the available information while rejecting other possibilities, such as ``Grasp the cosmetic.'' Operationally, we encode the image $I$ with an image encoder from CLIP \cite{clip} as a dense vector $I_e$. The binary tactile signal $T$ is transformed to a random initialized dense vector $T_e \in R^n$, with the same dimension as the embedding vectors in T5. Subsequently, image embedding $I_e$ and tactile embedding $T_e$ are concatenated with the word embedding $\Lambda_e$ of the given instruction $\Lambda$ as the input sequence $(\Lambda_e, T_e, I_e)$, as shown in Fig. \ref{fig:fg1}A. Given such input, the model is expected to generate the appropriate action language $C$ after training.

The network is trained with soft-prompt tuning \cite{prefix}. Conventional prompt tuning methods freeze all parameters of the pre-trained language model and use a language prompt to probe it to downstream tasks \cite{prefix}. In soft-prompt, the prompt is replaced by a group of trainable dense vectors, which avoids manually designing prompts and reduces the number of training parameters. Fig. \ref{fig:fg2}(b) shows the difference between our training method and alternatives. We add relevant soft prompts before the transformer layer of T5 to control the behavior of the LLM. Soft prompts are parameterized by using a two-layer feed-forward neural network. During training, we keep the language model parameters constant and only fine-tune the parameters related to these soft prompts. We learn the model with the language model loss: 
\begin{equation}
\label{loglike}
    L = -\sum^{m}_{i=1}  {\log P(c_{i}=c_{li}| c_{<i}, \Lambda,I,T)},
\end{equation}
where $m$ is the sequence length of $C$, $c_i \in V$ is the $i^{th}$ word in sequence $C$, $c_{li}$ is the $i^{th}$ word of sequence $C_l$, and $V$ is the vocabulary.

\subsection{Fine-resolution Inference}\label{sec:policy_rl}

The fine-resolution inference aims to link the action language inferred by the coarse-resolution module to the concrete action that enacts it. The fine resolution module outputs the action parameters of the gripper for the robot to execute, to realize the motor skills corresponding to the action language. These motor skills are limited to four degrees of freedom in order to simplify collision calculations and motion planning.

\textbf{Problem Formulation:} We formulate the problem of learning fine-resolution inference as a Markov Decision Process (MDP). An MDP comprises  state space $S^{\prime}$, action space $A$, a reward function $R(s_t, s_{t+1})$, and transition probability $P(s_{t+1} |s_t , a_t)$. The RL aims to discover an optimal policy $\pi $ that selects action $a_t$ to maximize cumulative rewards. 

\textbf{Learning Fine-resolution Inference:} We model the policy as a categorical model corresponding to a discrete-domain stochastic policy. The policy is trained with proximal policy optimization (PPO). At time step $t$, the agent chooses an action $a_t$ according to the probability output by the policy $\pi(a_t |s_t)$, and receives a reward $r_t$ from the environment.

The state is represented by a tuple $s_t = (I, L_e, T)$, where $I$ is the initial depth image of the environment with a resolution of 240×320, $T$ is a binary signal from the tactile sensor on the finger, and $L_e$ is the embedded vector of the inferred action language. $L_e$ and $T$ are concatenated as a vector $g_t = (L_e, T)$.

The policy network architecture comprises a convolutional (Conv) block and a multilayer perceptron (MLP) block, as shown in Fig. \ref{fig:fg1}B. The depth observation $I$ and $g_t$ are embedded into two latent vectors by the Conv block and MLP block, respectively. The resultant vectors are then concatenated and passed to the fully-connected layers (FCs) to produce an output action.

The action $a_t$ consists of two components, namely gripper pose displacement and gripper closure. The gripper pose displacement is constructed as the difference between the current and desired pose of the gripper. It is formed as $ (x_t,y_t,z_t,\alpha_t)$, where $(x_t,y_t,z_t)$ represents the relative displacement of the gripper in the workspace, and $\alpha_t$ represents the gripper's rotation about its z-axis. The displacement of the z-axis is executed last by the robot, and we fix the target gripper height during action execution to facilitate learning. The gripper closure control is represented by a one-hot vector $\beta_t$ is a one-hot vector that the gripper will be closed if $\beta_t = 1$. Thus, the full action is defined as $a_{t}=(x_t,y_t,z_t,\alpha_t,\beta_t)$, and we discretize each action coordinate according to the workspace. During operation, the robot initiates its movement along the x and y axes before proceeding to the z-axis.

The reward $r_t$ is given at the end of an episode, 1 for successfully completing the required motor skill and 0 otherwise. In addition, we also provide a linear auxiliary reward that encourages the robot to approach the target position. The auxiliary reward $r_{au}$ varies from 0 to 1 based on the distance between the gripper and target positions (the closer, the higher).

\begin{figure*}[!t]
\vspace{0.2cm}
    \centering
    \begin{overpic}[width=\linewidth]{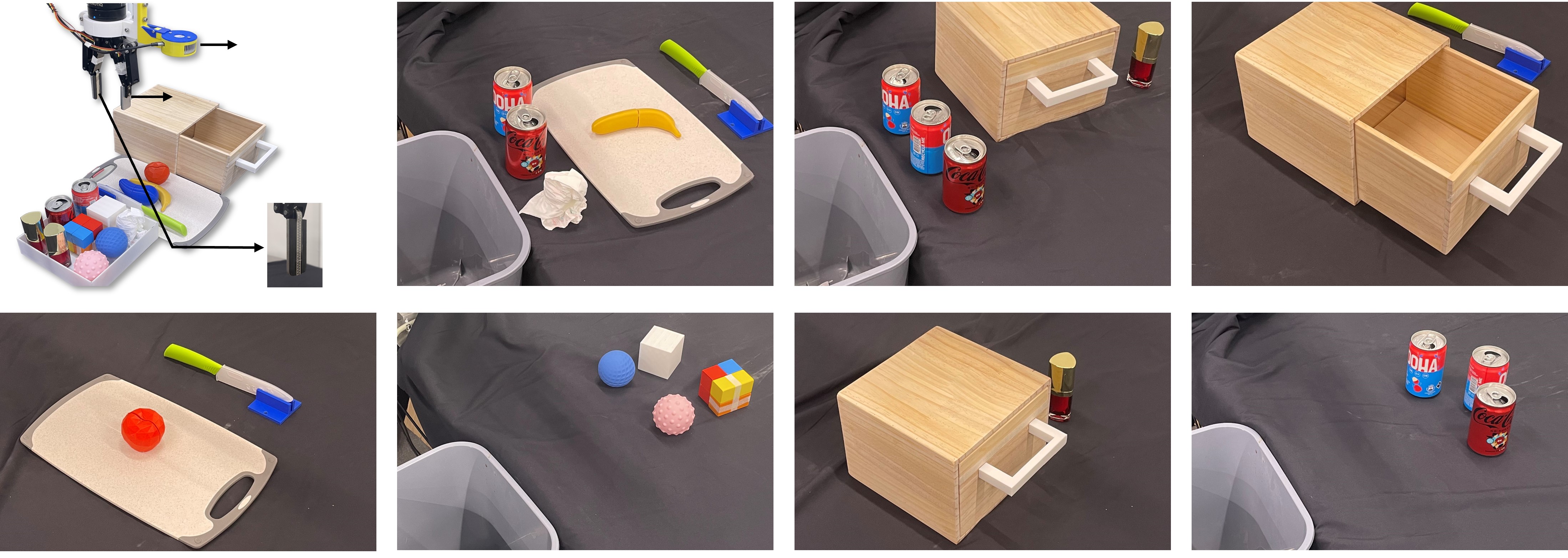}
    
    \put(15.25,32.25) {\footnotesize {Depth camera}}
    \put(11.25,29) {\footnotesize {Our gripper}}
    \put(15,23) {\footnotesize {Tactile sensor}}
    
    \put(11.25,16.2) {\footnotesize {(a)}}
    \put(61.5,16.1) {\footnotesize {(b)}}
    \put(49,-0.7) {\footnotesize {(c)}}

    \put(33, 18.75) {\scriptsize \textcolor{white}{Cut banana and clean table}}

    \put(59.5, 19.5) {\scriptsize \textcolor{white}{Put cosmetic into drawer}}
    \put(65, 18) {\scriptsize \textcolor{white}{and clean table}}

    \put(89, 19.5) {\scriptsize \textcolor{white}{Close drawer and}}
    \put(92.5, 18) {\scriptsize \textcolor{white}{grasp knife}}

    \put(17.5, 1.7) {\scriptsize \textcolor{white}{Cut apple}}

    \put(40.5, 2.5) {\scriptsize \textcolor{white}{Pick all round}}
    \put(39.25, 1) {\scriptsize \textcolor{white}{objects into box}}

    \put(17.5, 1.7) {\scriptsize \textcolor{white}{Cut apple}}
    \put(59.5, 1.7) {\scriptsize \textcolor{white}{Put cosmetic into drawer}}
    \put(92.5, 1.7) {\scriptsize \textcolor{white}{Clean table}}

    \end{overpic}
    \caption{ \textbf{Hardware and Scene Setup.} (a) Robot hardware and objects in real-world experiments; (b) Three scene setups for the hybrid tasks; (c) Four scene setups for the Long-horizon tasks. Short-horizon tasks also use these scenes.  An example task is shown at the bottom right in each scene setup.}
    \label{fig:fg4}
\vspace{-0.15cm}
\end{figure*}
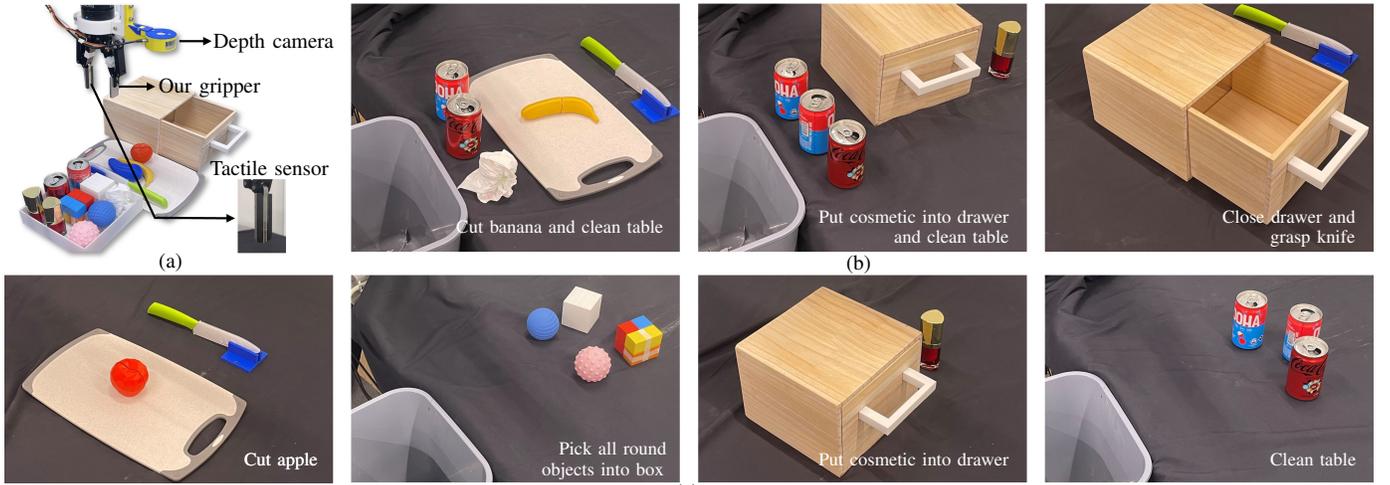

\begin{table*}[!t]
\caption{task family and language instruction definitions}
\centering
\resizebox{\linewidth}{!}{
\begin{tblr}{
  rowsep  = 1.5pt,
  cells = {c},
  row{3} = {t},
  row{5} = {t},
  row{7} = {t},
  cell{2}{1} = {r=2}{},
  cell{2}{2} = {r=2}{},
  cell{2}{3} = {b},
  cell{2}{4} = {r=2}{},
  cell{2}{5} = {r=2}{},
  cell{2}{6} = {r=2}{},
  cell{4}{1} = {r=2}{},
  cell{4}{2} = {r=2}{},
  cell{4}{3} = {b},
  cell{4}{4} = {r=2}{},
  cell{4}{5} = {r=2}{},
  cell{4}{6} = {b},
  cell{6}{1} = {r=2}{},
  cell{6}{2} = {r=2}{},
  cell{6}{3} = {b},
  cell{6}{4} = {r=2}{},
  cell{6}{5} = {b},
  cell{6}{6} = {b},
  hline{1-2,8} = {-}{},
}
Task Family & Num & Task Explanation & Instruction Type & Example Instruction & Example Task\\
Short-horizon & 10 & Tasks that require one reasoning step & Straight& ``Please grasp the apple'' & Robot needs to grasp the apple\\
 &  & completed by a single motor skill &  &  & \\
Long-horizon & 4 & Tasks that require many reasoning steps & Abstract& ``Please clean the table'' & Robot needs to pick up all trash\\
 &  & ~ ~ ~ ~completed by a range of motor skills~ ~ ~ ~ &  &  & on the table into the bin\\
Hybrid & 3 & Combined long-horizon & Abstract & ``Please put the apple into the & Robot needs to place the apple\\
 &  & and short-horizon tasks &  & drawer~and clean the table'' & and clean all trash
\end{tblr}
}
\vspace{-0.55cm}
\label{tab1}
\end{table*}

\subsection{Training Details}
Coarse-resolution and fine-resolution inferences are learned in the Pybullet simulator \cite{coumans2021} and then transferred to the real world. In the coarse-resolution inference learning stage, we generate 300 examples per task in the simulation for 17 language-conditional manipulation tasks, as shown in Fig. \ref{fig:fg5}. Specifically, we vary a task from the following three aspects: object types, object positions, and initial setups. For example, in the ``Put something into the drawer'' task, we consider different objects, such as cosmetics or cans, and randomly position them within the workspace. Additionally, we have two different initial setups where the drawer is either open or closed. The generated synthetic dataset $D$ contains over 12000 corresponding relationships and is used to train the coarse-resolution inference with cross-entropy loss. We follow the implementation in \cite{prefix} for the soft-prompt tuning. We use the Adam optimizer \cite{adam} and a linear learning rate scheduler during training. A default setting trains for ten epochs and uses a learning rate of $5\times 10^{-5}$. Fig. \ref{fig:fg4} shows seven scene setups, and we also create similar ones in the simulation. In the real world, we deploy ERRA on a UR10 arm equipped with a parallel gripper, an Intel L515 depth camera, and a tactile sensor, as shown in Fig. \ref{fig:fg4}(a). 

To learn the fine-resolution inference, 32 robots in simulation environments collect training episodes by obtaining the current policy from the optimizer every eight epochs. In each environment, a manipulation task specified by an action language is procedurally generated, which is randomly selected from substeps in 17 language-conditional manipulation tasks and applies the same random variations as during data collection in the coarse-resolution inference learning stage. The robot in the simulation environment then collects episodes, during which the reward is automatically determined based on whether the task is completed. If the robot completes the task, the environment will be reset, and a new task will be generated again. At last, the collected episodes are returned to the optimizer for learning the policy. During the training, we use Adam optimizer \cite{adam} with a learning rate of $10^{-4}$. We also randomize the object's physical properties during the task generation and add noise to the depth observation to make the learned policy robust to the various conditions in the real world. Specifically, the object size undergoes a global scaling, which entails resizing object dimensions within a range from 85\% (min) to 115\% (max) of its original size. Meanwhile, the spatial and visual properties are impacted by adding noise to camera properties. Specifically, we add noise to the camera position, camera pointing position, and field of view. The camera position and pointing are perturbed using three-dimensional vectors, and random noise of each dimension is sampled from a range $\{-2.5~mm, 2.5~mm\}$. Similarly, the field of view is perturbed with a noise range of $\{-0.025\degree, 0.025\degree\}$. Supplement materials are available at: {\color{blue}{\url{https://robotll.github.io/ERRA/}}}

\section{Experiments}\label{sec: exp_total}

We design a set of experiments in both simulation and real-world to evaluate the ERRA and other baselines in the language-conditioned manipulation tasks. The hypotheses we want t
o validate are as follows: 

\begin{itemize}
\item[\textit{H1:}] ERRA can perform long-horizon language-conditioned manipulation tasks and outperforms other baselines. 
\item[\textit{H2:}] Robot proprioception is important for completing language-conditioned manipulation tasks. 
\item[\textit{H3:}] LLMs with prompt-tuning allow ERRA to generalize to unseen natural language instructions.  
\item[\textit{H4:}] ERRA is able to transfer to the real world.
\item[\textit{H5:}] ERRA can respond to environmental changes caused by humans or its own failures. 
\end{itemize}

\subsection{Scenes, Tasks and Evaluation Setup}

\textbf{Scene and task setup:} Fig. \ref{fig:fg4} shows seven scene setups, and we also create similar ones in the simulation. Our hardware settings in the real world are also shown in Fig. \ref{fig:fg4}(a). To evaluate ERRA, we test its performance on 17 language-conditioned manipulation tasks from seven scenes in both simulation and real-world. These tasks cover time horizons, language complexity, and variations over the robot and environment. Tab. \ref{tab1} details examples for each task family, which fall into the following: 


\begin{itemize}

\item \textbf{Short-horizon}: Short-horizon tasks are decomposed from long-horizon tasks, which involve a straight language instruction that needs to be achieved by a single motor skill. The instruction and the action language have a one-to-one correspondence in such tasks. 

\item \textbf{Long-horizon:} Tasks are specified by abstract natural language instruction and achieved by a long sequence of motor skills. The correspondence between the language instruction and the action language is not one-to-one and is affected by the environment and robot state. This tests the ERRA's ability to reason abstract instructions and to plan with the environment's causal relation.

\item \textbf{Hybrid:} These tasks are the combination of multiple long-horizon and short-horizon tasks, which have a higher complexity than others.

\end{itemize}

\textbf{Baseline comparisons:} We compare with the following approaches:
\begin{itemize}
\item \textbf{Infer-all:} It is similar to the architecture of SayCan \cite{saycan}, in which all action languages are inferred together, and then the robot executes them one by one without feedback during the entire task planning and execution process.
\item \textbf{ERRA-w/o touch:} An ablated version of the ERRA without the proprioceptive input (i.e.,  tactile information). Both coarse-resolution and fine-resolution modules only use the camera to observe environments.
\item \textbf{ERRA:}  We deploy the ERRA system to the robot, which is the full non-ablated method we propose in this article.
\end{itemize}

\begin{figure}[!t]
    \centering
    \begin{overpic}[width=\linewidth]{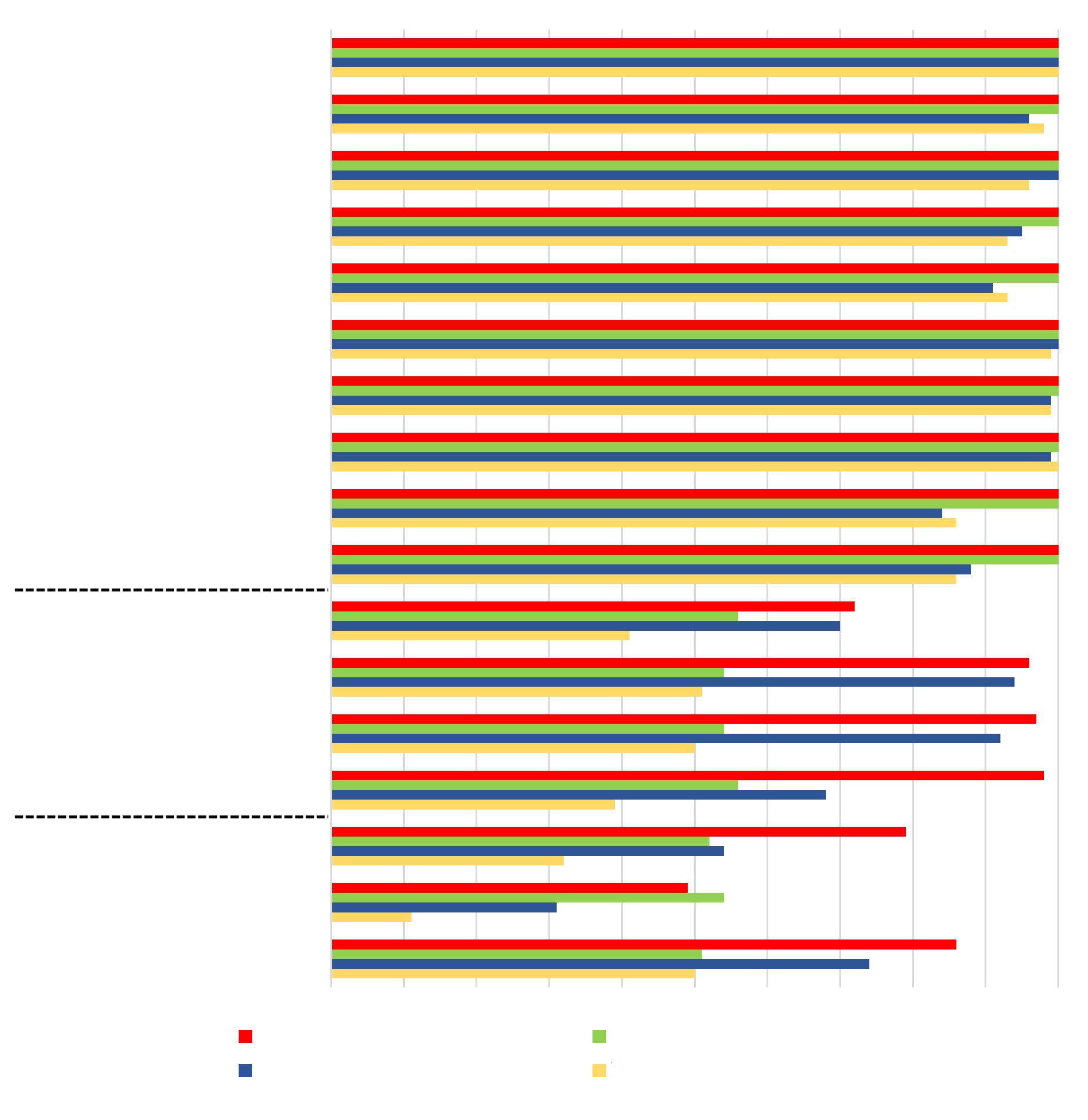}
    
    \put(9.2,94.4) {\tiny {Place the round object}}
    \put(10.2,89.4) {\tiny {Grasp a round object}}
    \put(10,84.4) {\tiny {Place sth into the bin}}
    \put(9.25,79.4) {\tiny {Grasp sth on the table}}
    \put(12.5,74.4) {\tiny {Cut sth with knife}}
    \put(16,69.4) {\tiny {Pick the knife}}
    \put(13.5,64.4) {\tiny {Close the drawer}}
    \put(7,59.4) {\tiny {Place sth into the drawer}}
    \put(6.5,54.4) {\tiny {Grasp sth near the drawer}}
    \put(14,49.5) {\tiny {Open the drawer}}

    \put(9.9,44.3) {\tiny {Pick all round objects}}
    \put(15.1,39.3) {\tiny {Clean the table}}
    \put(7.7,34.4) {\tiny {Cut sth (need find knife)}}
    \put(8.9,29.4) {\tiny {Put sth into the drawer}}

    \put(8.8,24) {\tiny {Clean table and cut sth}}
    \put(7.4,18.9) {\tiny {Place sth and clean table}}
    \put(5.6,13.9) {\tiny {Close drawer and grasp sth}}

    \put(29,10.25) {\tiny {0\%}}
    \put(41,10.25) {\tiny {20\%}}
    \put(54,10.25) {\tiny {40\%}}
    \put(67,10.25) {\tiny {60\%}}
    \put(80,10.25) {\tiny {80\%}}
    \put(92.5,10.25) {\tiny {100\%}}

    \put(1, 65){ {\rotatebox{90}{\scriptsize Short-horizon}}}
    \put(1, 29.5){ {\rotatebox{90}{\scriptsize Long-horizon}}}
    \put(1, 16){ {\rotatebox{90}{\scriptsize Hybird}}}

    \put(23,6.8) {\tiny {Plan success rate (ERRA)}}
    \put(23,3.8) {\tiny {Task success rate (ERRA)}}

    \put(54.75,6.8) {\tiny {Plan success rate (Infer-all)}}
    \put(54.75,3.8) {\tiny {Task success rate (Infer-all)}}
    
    \end{overpic}
    \vspace*{-7mm}
    \caption{\textbf{Task performance in the simulation.} From top to bottom, there are 14 short-horizon tasks, four long-horizon tasks, and three hybrid tasks.
    }
    \label{fig:fg5}
\vspace{-0.1cm}
\end{figure}
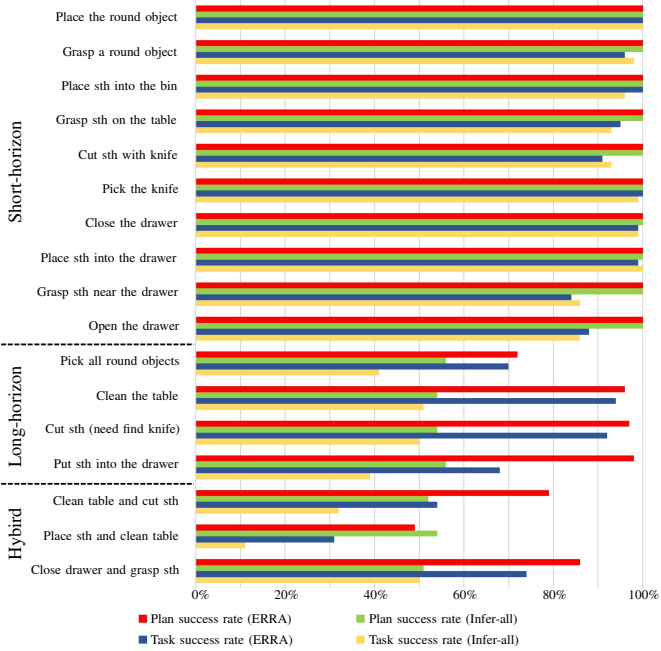

\begin{table}[!t]
\caption{simulation experiments}
\centering
\begin{threeparttable}
\resizebox{\linewidth}{!}{
\begin{tblr}{
  cells = {c},
  cell{1}{1} = {r=2}{},
  cell{1}{2} = {c=2}{},
  cell{1}{4} = {c=2}{},
  cell{1}{6} = {c=2}{},
  cell{1}{8} = {c=2}{},
  hline{1,6} = {-}{},
  hline{3} = {2-9}{},
}
Method & Short-horizon &  & Long-horizon &  & Hybrid &  & Total & \\
 & Plan\textsuperscript{*} & Task\textsuperscript{**} & Plan & Task & Plan & Task & Plan & Task\\
Infer-all & \textbf{100\%} &\textbf{ 94\%} & 55\% & 46\% & 52\% & 31\% & 69\% & 57\% \\
ERRA-w/o touch & 83\% & 79\% & 48\% & 41\% & 42\% & 35\% & 58\% & 52\% \\
\textbf{ERRA} & \textbf{100\%} & \textbf{94\%} & \textbf{91\%} & \textbf{81\%} & \textbf{77\%} & \textbf{64\%} & \textbf{89\% }& \textbf{80\%}
\end{tblr}
}
\begin{tablenotes}
\item[*]\textit{Plan success rate.} $^{**}$\textit{Task success rate.}
\end{tablenotes}
\end{threeparttable}
\label{tab2}
\vspace{-0.5cm}
\end{table}

\textbf{Metric:} We consider two evaluation metrics: \textit{plan success rate} (successful task planning/total attempts) and \textit{task success rate} (completed tasks/total attempts) for validating performance. The \textit{plan success rate} is measured by whether the module of coarse-resolution inference correctly predicts all action languages in a language-conditioned manipulation task, assuming that the execution of motor skills is flawless. The \textit{task success rate} is calculated based on whether the target manipulation task is completed. It requires the coarse-resolution module to successfully plan each step and the fine-resolution module to output the correct actions for the robot to complete corresponding motor skills. For each task, we repeat the test 500 times in simulation experiments and ten times in real-world experiments.

\begin{figure}[!t]
\vspace{0.4cm}
    \centering
    
    \begin{overpic}[width=\linewidth]{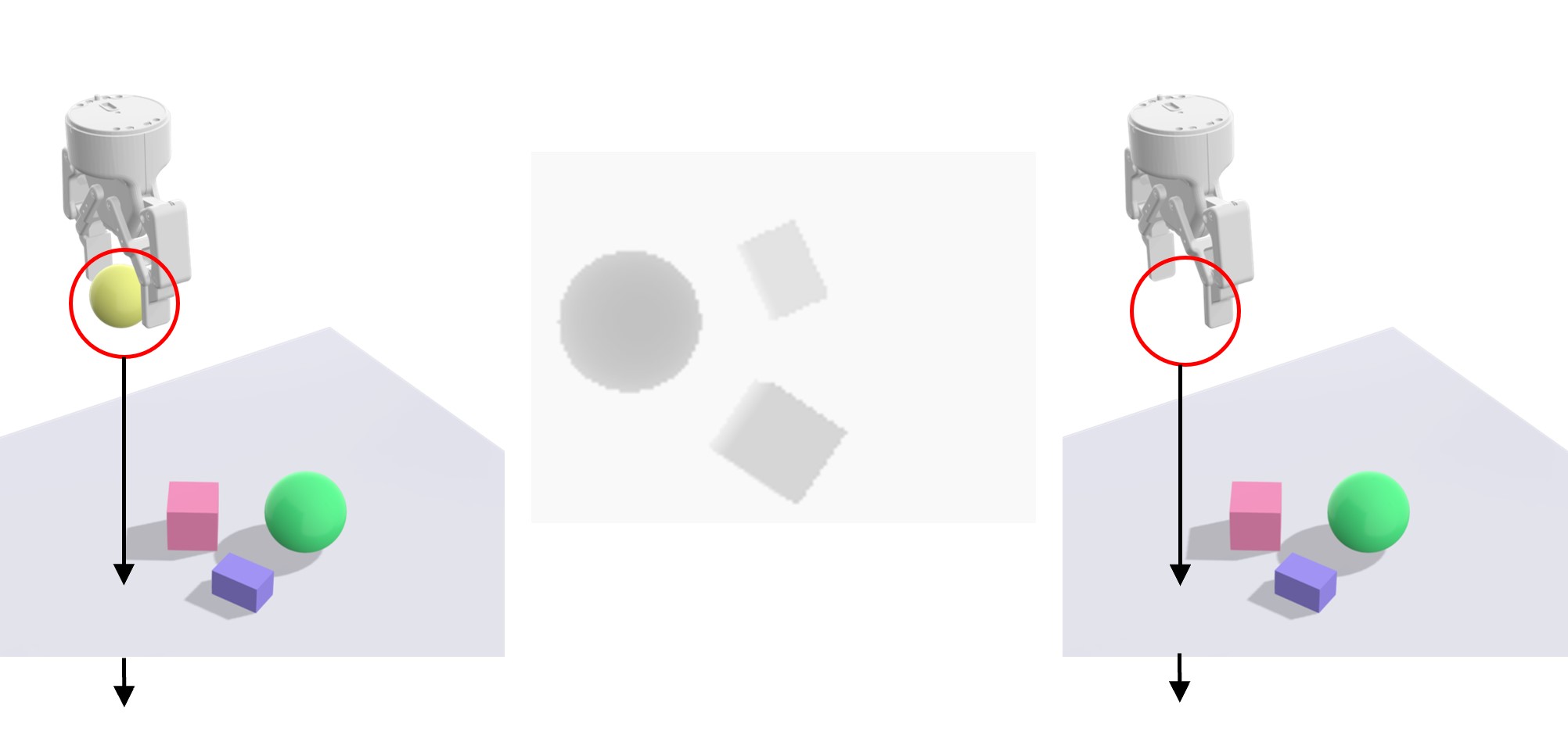}
    \put(23,44) {\footnotesize {Same instruction: Please clean the table}}
    \put(33,40) {\footnotesize {Same camera observation}}

    \put(2,6.6) {\footnotesize {Tactile: 1}}
    \put(1,-0.25) {\footnotesize {Next: Put the object into the bin}}

    \put(69,6.6) {\footnotesize {Tactile: 0}}
    \put(69,-0.25) {\footnotesize {Next: Grasp an object}}

    \end{overpic}
    \caption{A case where two scenes have the same instructions and visual observations but different next plans. Language and visual information are insufficient to plan the next step without tactile signals.}
    \label{fig:fg3}
\vspace{-0.25cm}
\end{figure}

\subsection{Simulation Results}\label{sec: simulation results}

\textbf{Comparison to baselines:} Tab. \ref{tab2} shows the performance of ERRA on different task families in the simulation. Across all tasks, ERRA achieves a plan success rate of 89\% and a task success rate of 80\%. In the Long-horizon and Hybrid families, ERRA achieves 91\% and 77\% plan success rates, respectively. Such results highlight the effectiveness of coarse-resolution inference in enabling ERRA to reason and plan for tasks with longer horizons. The plan and task success rate for each task is fully illustrated in Fig. \ref{fig:fg5}. 

To demonstrate the importance of incorporating robot proprioception, we conduct an ablation experiment by excluding the tactile information from inputs during ERRA training (denoted as ERRA-w/o touch in Tab. \ref{tab2}). The results show that the planning performance of ERRA -w/o touch is reduced by up to 43\% on the Long-horizon family and 35\% on the Hybrid family. This decline is attributed to the incomplete information required for reasoning. In certain tasks, the relationship between state and action language is not one-to-one, thereby rendering reasoning impossible. An example of such a scenario is presented in Fig. 4.


We then investigate the effectiveness of the coarse-to-fine inference design in the ERRA by comparing the ERRA with an architecture in which the planning and execution are independent (denoted as infer-all in Tab. \ref{tab2}). Our results reveal that ERRA outperforms the Infer-all by over 25\% on the Long-horizon and Hybrid families. Infer-all's suboptimal performance is due to its reliance on accurately inferring all action language before executing corresponding motor skills, which increases the difficulty of reasoning and planning. In contrast, the ERRA utilizes closed-loop feedback by inferring the next step only after the robot executes the previous step, leading to more effective and robust task performance. 

\begin{figure*}[!t]
    \centering
    \begin{overpic}[width=\linewidth]{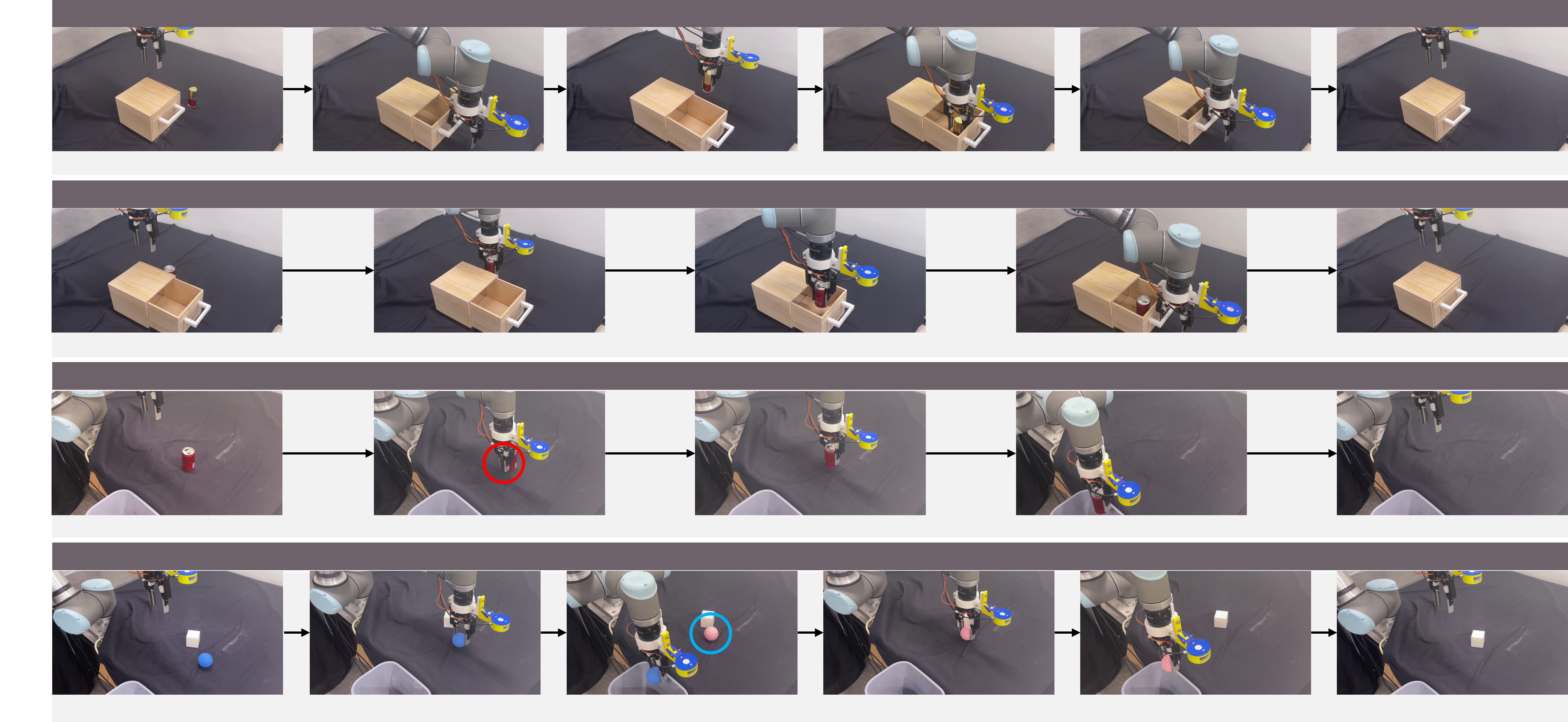}

    \put(4,44.75) {\scriptsize \textcolor{white}{Instruction: ``Please put the cosmetic into the drawer''}}
    \put(4,33.25) {\scriptsize \textcolor{white}{Instruction: ``Please put the can into the drawer''}}
    \put(4,21.7) {\scriptsize \textcolor{white}{Instruction: ``Please put clean the table''}}
    \put(4,10.1) {\scriptsize \textcolor{white}{Instruction: ``Please put all round objects into the bin''}}
    
    \put(0.6,40.25) {\scriptsize {A:}}
    \put(0.6,17.25) {\scriptsize {B:}}
    \put(0.6,5.75) {\scriptsize {C:}}

    \put(4,35.25) {\scriptsize {Action language:}}
    \put(22,35.25) {\scriptsize {Open the drawer}}
    \put(38,35.25) {\scriptsize {Grasp the cosmetic}}
    \put(51,35.25) {\scriptsize {Put the cosmetic into the drawer}}
    \put(71.5,35.25) {\scriptsize {Close the drawer}}
    \put(91,35.25) {\scriptsize {Done}}

    \put(4,23.7) {\scriptsize {Action language:}}
    \put(27,23.7) {\scriptsize {Grasp the can}}
    \put(44,23.7) {\scriptsize {Put the can into the drawer}}
    \put(67.5,23.7) {\scriptsize {Close the drawer}}
    \put(91,23.7) {\scriptsize {Done}}

    \put(4,12.05) {\scriptsize {Action language:}}
    \put(29.5,14.5) {\tiny \textcolor{red}{Object slipped}}
    \put(27,12.05) {\scriptsize {Grasp an object}}
    \put(47,12.05) {\scriptsize {Grasp an object}}
    \put(64.8,12.05) {\scriptsize {Put the object into the bin}}
    \put(91,12.05) {\scriptsize {Done}}

    \put(4,0.5) {\scriptsize {Action language:}}
    \put(43, 7.25) {\tiny \textcolor{cyan}{Human added}}
    \put(21,0.5) {\scriptsize {Grasp a round object}}
    \put(34.6,0.5) {\scriptsize {Put the round object in the bin}}
    \put(54,0.5) {\scriptsize {Grasp a round object}}
    \put(67.5,0.5) {\scriptsize {Put the round object in the bin}}
    \put(91,0.5) {\scriptsize {Done}}

    \end{overpic}
    \caption{\textbf{Qualitative results of ERRA.}
    \textbf{A:} Sequences of the robot successfully placing the object into the drawer at different settings (i.e., drawer is closed or open);
    \textbf{B:} A sequence of the robot successfully recovering from its execution failure (i.e., object slip from gripper) and complete table cleaning;
   \textbf{C:} A sequence of the robot adapting to the dynamic environment. (i.e., human places another round object).
    }
    \label{fig:fg6}
\vspace{-0.3cm}
\end{figure*}
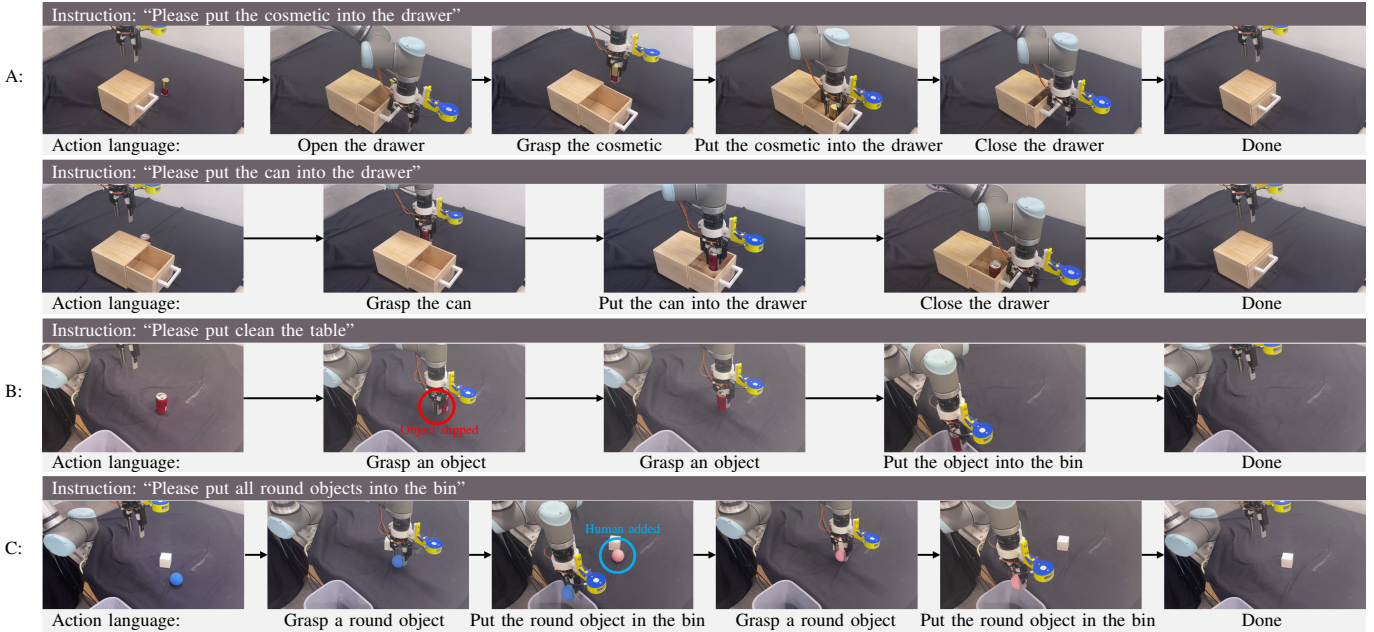

\begin{table}[!t]
\centering
\vspace{0.25cm}
\caption{generalization to unseen language instructions}
\resizebox{\linewidth}{!}{
\begin{tblr}{
  cells = {c},
  cell{1}{1} = {r=2}{},
  cell{1}{2} = {c=2}{},
  cell{1}{4} = {c=2}{},
  cell{1}{6} = {c=2}{},
  cell{1}{8} = {c=2}{},
  hline{1,6} = {-}{},
  hline{3} = {2-9}{},
}
Type & Short-horizon &  & Long-horizon &  & Hybrid &  & Total & \\
 & Plan & Task & Plan & Task & Plan & Task & Plan & Task\\
Unseen Verb & 99\% & 94\% & 77\% & 68\% & 51\% & 42\% & 76\% & 68\%\\
Unseen Noun & 100\% & 94\% & 80\% & 72\% & 55\% & 40\% & 78\% & 69\%\\
Unseen Verb + Noun & 99\% & 94\% & 52\% & 47\% & 34\% & 25\% & 62\% & 55\%
\end{tblr}
}
\label{tab3}
\vspace{-0.5cm}
\end{table}

\textbf{Generalization to unseen language instructions:} We study the ERRA's generalization ability to unseen natural language instructions. Specifically, we test the generalization of ERRA at three levels of rephrased language instructions with novel but similar words. First, we replace nouns in the language instructions of tasks (e.g., ``cut the banana''  to ``chop the banana''), denoted as Unseen Verb in Tab. \ref{tab3}. Second, we replace verbs (e.g., ``grasp the cola''  to ``grasp the can''), denoted as Unseen Noun in Tab. \ref{tab3}. Finally,  we replace both nouns and verbs in instructions(e.g., ``close the drawer'' to ``shut the cabinet''), denoted as Unseen Verb + Noun in Tab. \ref{tab3}. 

{As shown in Tab. \ref{tab3}, ERRA's generalization performance degrades as the complexity of the task and the number of unseen words in the instruction increase. Specifically, in the Long-horizon family, changing either the verbs or nouns leads to a 12\% decrease in plan success rate, while changing both results in a 40\% decline. The decline in planning performance also leads to a corresponding decrease in task success rate. In contrast, ERRA remains stable in the Short-horizon family, owing to the lower complexity of language abstraction and task planning. Notably, we observe that the performance of the Hybrid family drops by up to 35\% on novel instructions due to the challenging language instructions and task planning involved. In conclusion, our findings suggest that the ERRA can generalize to novel language instructions, but its generalization performance is affected by the complexity of the task and the number of unseen words in the instruction.


\subsection{Real-world Experiments}\label{sec:real_robot_exp}


We also evaluate ERRA's performance in the real world. ERRA achieves an average task success rate of 77\% across three task families. The results show that the model's reasoning, planning, and interaction abilities are transferable to real-world scenes for solving long-horizon language-conditioned manipulation tasks. Similar to the simulation results, ERRA performs best (89\% success rate) on Short-horizon tasks among the three task families. The performance of ERRA decreases as task complexity increases, achieving a success rate of 75\%  on the Long-horizon family and 68\% on the Hybrid family. The running time for one step in the task is approximately 12 seconds, which includes the entire cycle time from network inferences (less than 0.2~s) to robot execution. 

\begin{table}[!t]
\caption{real world results}
\centering
\resizebox{\linewidth}{!}{
\begin{tblr}{
  row{1} = {c},
  cell{1}{1} = {r=2}{},
  cell{1}{2} = {c=4}{},
  cell{2}{2} = {c},
  cell{3}{1} = {c},
  cell{3}{2} = {c},
  cell{3}{3} = {c},
  cell{3}{4} = {c},
  hline{1,4} = {-}{},
  hline{3} = {2-5}{},
}
Method & Task Success Rate &  &  & \\
 & Short-horizon & Long-horizon & Hybrid & Total\\
ERRA & 89\% & 75\% & 68\% & 77\%
\end{tblr}
}
\label{tab4}
\vspace{-0.5cm}
\end{table}

Looking back to our initial example in Sec. \ref{intro}, ``Please put the cosmetic in the drawer,'' we have demonstrated that ERRA is able to discover whether the robot needs ``Open the drawer'' by reasoning the causal relation of the environment  (See Fig. \ref{fig:fg6}A) and plan and execute a long sequence in the real world, which include opening the drawer, grasping the cosmetic, putting the cosmetic into the drawer and then closing the drawer. Note also that the robot only has one arm, ERRA necessitates planning the action in reasonable order (e.g., first, open the drawer and then grasp the cosmetic, not the other way around). This requires the ERRA to have strong abilities of long-horizon reasoning and understanding of semantic knowledge in the language instruction.

As shown in Fig. \ref{fig:fg6}, ERRA manifests robustness to dynamic environments. ERRA discovers a new round object added by the human after the last object has been put in the bin and correctly infers that the next action language is ``grasp a round object'' rather than ``Done'' (See Fig. \ref{fig:fg6}C). Such behavior is powered by itself, benefiting from the closed-loop feedback provided by the coarse-to-fine inference architecture. Such feedback also allows the robot interactively recover from failure cases. Fig. \ref{fig:fg6}B shows the ERRA response to its failure (object slip from hand during the task execution).

\section{Conclusion, Limitation, and Future Work}

We have presented a novel solution, ERRA, that utilizes tightly-coupled probabilistic inferences at two granularity levels, coarse and fine, for solving long-horizon language-conditioned manipulation tasks. Through coarse-to-fine inferences, complex manipulation tasks can be decomposed into concrete actions and executed by the robot. Extensive controlled experiments demonstrate the robustness and effectiveness of ERRA on manipulation tasks with long-horizon and abstract semantics. Our work is not without limitations; first, limited by hardware devices, the robot`s position is fixed without the need for localization and mapping, suggesting exciting opportunities for extending the current work to the scene of mobile robots. Future research can also benefit from the flexibility and efficiency of a dual-arm system. While planning for the dual-arm system may involve evaluating a larger number of potential actions, the increased flexibility and redundancy provided by the additional arm may result in more optimal final plans, compared to our single-arm system. Finally, the proposed work relies on simulated data to learn inference at both coarse and fine resolutions, which is a significant advantage that avoids a more time-consuming process, such as manual labeling. However, it still needs to build simulation scenes carefully. One possible opportunity is to develop a method that is able to learn from online videos in which humans perform manipulation tasks with long-term and abstract semantics.

\normalem
\bibliographystyle{ieeetr}
\bibliography{references}

\begin{thebibliography}{10}

\bibitem{saycan}
M.~Ahn, A.~Brohan, N.~Brown, Y.~Chebotar, O.~Cortes, B.~David, C.~Finn,
  K.~Gopalakrishnan, K.~Hausman, A.~Herzog, {\em et~al.}, ``Do as i can, not as
  i say: Grounding language in robotic affordances,'' in {\em 6th Annual
  Conference on Robot Learning}, 2022.

\bibitem{inner}
W.~Huang, F.~Xia, T.~Xiao, H.~Chan, J.~Liang, P.~Florence, A.~Zeng, J.~Tompson,
  I.~Mordatch, Y.~Chebotar, P.~Sermanet, T.~Jackson, N.~Brown, L.~Luu,
  S.~Levine, K.~Hausman, and brian ichter, ``Inner monologue: Embodied
  reasoning through planning with language models,'' in {\em 6th Annual
  Conference on Robot Learning}, 2022.

\bibitem{huang2022language}
W.~Huang, P.~Abbeel, D.~Pathak, and I.~Mordatch, ``Language models as zero-shot
  planners: Extracting actionable knowledge for embodied agents,'' {\em arXiv
  preprint arXiv:2201.07207}, 2022.

\bibitem{akbari2019knowledge}
A.~Akbari, Muhayyuddin, and J.~Rosell, ``Knowledge-oriented task and motion
  planning for multiple mobile robots,'' {\em Journal of Experimental \&
  Theoretical Artificial Intelligence}, vol.~31, no.~1, pp.~137--162, 2019.

\bibitem{sy1}
M.~Sridharan, M.~Gelfond, S.~Zhang, and J.~Wyatt, ``Reba: A refinement-based
  architecture for knowledge representation and reasoning in robotics,'' {\em
  Journal of Artificial Intelligence Research}, vol.~65, pp.~87--180, 2019.

\bibitem{op1}
M.~A. Toussaint, K.~R. Allen, K.~A. Smith, and J.~B. Tenenbaum,
  ``Differentiable physics and stable modes for tool-use and manipulation
  planning,'' Robotics: Science and Systems Foundation, 2018.

\bibitem{llm1}
Y.~Jiang, A.~Gupta, Z.~Zhang, G.~Wang, Y.~Dou, Y.~Chen, L.~Fei-Fei,
  A.~Anandkumar, Y.~Zhu, and L.~Fan, ``Vima: General robot manipulation with
  multimodal prompts,'' {\em arXiv preprint arXiv:2210.03094}, 2022.

\bibitem{llm2}
J.~Wei, Y.~Tay, R.~Bommasani, C.~Raffel, B.~Zoph, S.~Borgeaud, D.~Yogatama,
  M.~Bosma, D.~Zhou, D.~Metzler, E.~H. Chi, T.~Hashimoto, O.~Vinyals, P.~Liang,
  J.~Dean, and W.~Fedus, ``Emergent abilities of large language models,'' {\em
  Transactions on Machine Learning Research}, 2022.
\newblock Survey Certification.

\bibitem{llm3}
D.~Shah, B.~Osi{\'n}ski, S.~Levine, {\em et~al.}, ``Robotic navigation with
  large pre-trained models of language, vision, and action,'' in {\em 6th
  Annual Conference on Robot Learning}.

\bibitem{llm4}
A.~Ramesh, P.~Dhariwal, A.~Nichol, C.~Chu, and M.~Chen, ``Hierarchical
  text-conditional image generation with clip latents,'' {\em arXiv preprint
  arXiv:2204.06125}, 2022.

\bibitem{prefix}
X.~L. Li and P.~Liang, ``Prefix-tuning: Optimizing continuous prompts for
  generation,'' in {\em Proceedings of the 59th Annual Meeting of the
  Association for Computational Linguistics and the 11th International Joint
  Conference on Natural Language Processing (Volume 1: Long Papers)},
  pp.~4582--4597, 2021.

\bibitem{lcmf-1}
Y.~Chen, R.~Xu, Y.~Lin, and P.~A. Vela, ``A joint network for grasp detection
  conditioned on natural language commands,'' in {\em 2021 IEEE International
  Conference on Robotics and Automation (ICRA)}, pp.~4576--4582, IEEE, 2021.

\bibitem{lcmf-2}
S.~Stepputtis, J.~Campbell, M.~Phielipp, S.~Lee, C.~Baral, and H.~Ben~Amor,
  ``Language-conditioned imitation learning for robot manipulation tasks,''
  {\em Advances in Neural Information Processing Systems}, vol.~33,
  pp.~13139--13150, 2020.

\bibitem{lcmf-3}
C.~Wang, C.~Ross, Y.-L. Kuo, B.~Katz, and A.~Barbu, ``Learning a
  natural-language to ltl executable semantic parser for grounded robotics,''
  in {\em Conference on Robot Learning}, pp.~1706--1718, PMLR, 2021.

\bibitem{lcmf-4}
V.~Blukis, R.~Knepper, and Y.~Artzi, ``Few-shot object grounding and mapping
  for natural language robot instruction following,'' in {\em Conference on
  Robot Learning}, pp.~1829--1854, PMLR, 2021.

\bibitem{lcmf-5}
W.~Liu, C.~Paxton, T.~Hermans, and D.~Fox, ``Structformer: Learning spatial
  structure for language-guided semantic rearrangement of novel objects,'' in
  {\em 2022 International Conference on Robotics and Automation (ICRA)},
  pp.~6322--6329, IEEE, 2022.

\bibitem{lcm-1-bcz}
E.~Jang, A.~Irpan, M.~Khansari, D.~Kappler, F.~Ebert, C.~Lynch, S.~Levine, and
  C.~Finn, ``Bc-z: Zero-shot task generalization with robotic imitation
  learning,'' in {\em Conference on Robot Learning}, pp.~991--1002, PMLR, 2022.

\bibitem{lcm-2}
L.~Shao, T.~Migimatsu, Q.~Zhang, K.~Yang, and J.~Bohg, ``Concept2robot:
  Learning manipulation concepts from instructions and human demonstrations,''
  {\em The International Journal of Robotics Research}, vol.~40, no.~12-14,
  pp.~1419--1434, 2021.

\bibitem{lcm-3}
S.~Stepputtis, J.~Campbell, M.~Phielipp, S.~Lee, C.~Baral, and H.~Ben~Amor,
  ``Language-conditioned imitation learning for robot manipulation tasks,''
  {\em Advances in Neural Information Processing Systems}, vol.~33,
  pp.~13139--13150, 2020.

\bibitem{lcm-e1}
M.~Shridhar, L.~Manuelli, and D.~Fox, ``Perceiver-actor: A multi-task
  transformer for robotic manipulation,'' {\em arXiv preprint
  arXiv:2209.05451}, 2022.

\bibitem{lcm-e2}
S.~Nair, E.~Mitchell, K.~Chen, S.~Savarese, C.~Finn, {\em et~al.}, ``Learning
  language-conditioned robot behavior from offline data and crowd-sourced
  annotation,'' in {\em Conference on Robot Learning}, pp.~1303--1315, PMLR,
  2022.

\bibitem{gozero}
D.~Silver, T.~Hubert, J.~Schrittwieser, I.~Antonoglou, M.~Lai, A.~Guez,
  M.~Lanctot, L.~Sifre, D.~Kumaran, T.~Graepel, {\em et~al.}, ``A general
  reinforcement learning algorithm that masters chess, shogi, and go through
  self-play,'' {\em Science}, vol.~362, no.~6419, pp.~1140--1144, 2018.

\bibitem{atari}
V.~Mnih, K.~Kavukcuoglu, D.~Silver, A.~A. Rusu, J.~Veness, M.~G. Bellemare,
  A.~Graves, M.~Riedmiller, A.~K. Fidjeland, G.~Ostrovski, {\em et~al.},
  ``Human-level control through deep reinforcement learning,'' {\em nature},
  vol.~518, no.~7540, pp.~529--533, 2015.

\bibitem{rl2}
H.-G. Cao, W.~Zeng, and I.-C. Wu, ``Reinforcement learning for picking
  cluttered general objects with dense object descriptors,'' in {\em 2022
  International Conference on Robotics and Automation (ICRA)}, pp.~6358--6364,
  IEEE, 2022.

\bibitem{rl3}
C.~Zhao, Z.~Tong, J.~Rojas, and J.~Seo, ``Learning to pick by digging:
  Data-driven dig-grasping for bin picking from clutter,'' in {\em 2022
  International Conference on Robotics and Automation (ICRA)}, pp.~749--754,
  IEEE, 2022.

\bibitem{rl4}
C.~Zhao and J.~Seo, ``Learn from interaction: Learning to pick via
  reinforcement learning in challenging clutter,'' in {\em 2022 IEEE/RSJ
  International Conference on Intelligent Robots and Systems (IROS)}, Oct.
  2022.

\bibitem{hrl1}
D.~Hafner, K.-H. Lee, I.~Fischer, and P.~Abbeel, ``Deep hierarchical planning
  from pixels,'' in {\em Advances in Neural Information Processing Systems}
  (A.~H. Oh, A.~Agarwal, D.~Belgrave, and K.~Cho, eds.), 2022.

\bibitem{hrl2}
F.~Xia, C.~Li, R.~Mart{\'\i}n-Mart{\'\i}n, O.~Litany, A.~Toshev, and
  S.~Savarese, ``Relmogen: Integrating motion generation in reinforcement
  learning for mobile manipulation,'' in {\em 2021 IEEE International
  Conference on Robotics and Automation (ICRA)}, pp.~4583--4590, IEEE, 2021.

\bibitem{t5}
C.~Raffel, N.~Shazeer, A.~Roberts, K.~Lee, S.~Narang, M.~Matena, Y.~Zhou,
  W.~Li, P.~J. Liu, {\em et~al.}, ``Exploring the limits of transfer learning
  with a unified text-to-text transformer.,'' {\em J. Mach. Learn. Res.},
  vol.~21, no.~140, pp.~1--67, 2020.

\bibitem{clip}
A.~Radford, J.~W. Kim, C.~Hallacy, A.~Ramesh, G.~Goh, S.~Agarwal, G.~Sastry,
  A.~Askell, P.~Mishkin, J.~Clark, {\em et~al.}, ``Learning transferable visual
  models from natural language supervision,'' in {\em International Conference
  on Machine Learning}, pp.~8748--8763, PMLR, 2021.

\bibitem{coumans2021}
E.~Coumans and Y.~Bai, ``Pybullet, a python module for physics simulation for
  games, robotics and machine learning.'' \url{http://pybullet.org},
  2016--2021.

\bibitem{adam}
D.~P. Kingma and J.~Ba, ``Adam: {A} method for stochastic optimization,'' in
  {\em 3rd International Conference on Learning Representations, {ICLR} 2015,
  San Diego, CA, USA, May 7-9, 2015, Conference Track Proceedings} (Y.~Bengio
  and Y.~LeCun, eds.), 2015.

\end{thebibliography}

\end{document}